\documentclass[10pt, a4paper]{article}

\usepackage{lrec-coling2024} 

\usepackage{multirow}
\usepackage{graphicx}
\usepackage{CJKutf8}
\usepackage{subcaption}
\usepackage{marvosym}

\title{Advancing Topic Segmentation and Outline Generation in Chinese Texts: 
The Paragraph-level Topic Representation, Corpus, and Benchmark}

\name{
\begin{tabular}{c}
Feng Jiang\textsuperscript{1,2}, Weihao Liu\textsuperscript{3}, Xiaomin Chu\textsuperscript{3}, \\
Peifeng Li\textsuperscript{3}, Qiaoming Zhu\textsuperscript{3}, Haizhou Li\textsuperscript{1}\sthanks{ \quad Corresponding Author.}
\end{tabular}
} 

\address{\textsuperscript{1}Shenzhen Research Institute of Big Data, School of Data Science, \\ The Chinese University of Hong Kong, Shenzhen (CUHK-Shenzhen), Guangdong, China \\
        \textsuperscript{2}School of Information Science and Technology,\\
        University of Science and Technology of China, Hefei, China \\
        \textsuperscript{3}School of Computer Science and Technology, Soochow University, Suzhou, China \\
         \{jeffreyjiang,haizhouli\}@cuhk.edu.cn,\\
         whliu@stu.suda.edu.cn, \{xmchu, pfli, qmzhu\}@suda.edu.cn\\}

\abstract{
Topic segmentation and outline generation strive to divide a document into coherent topic sections and generate corresponding subheadings, unveiling the discourse topic structure of a document. Compared with sentence-level topic structure, the paragraph-level topic structure can quickly grasp and understand the overall context of the document from a higher level, benefitting many downstream tasks such as summarization, discourse parsing, and information retrieval. However, the lack of large-scale, high-quality Chinese paragraph-level topic structure corpora restrained relative research and applications. To fill this gap, we build the Chinese paragraph-level topic representation, corpus, and benchmark in this paper. Firstly, we propose a hierarchical paragraph-level topic structure representation with three layers to guide the corpus construction. Then, we employ a two-stage man-machine collaborative annotation method to construct the largest Chinese Paragraph-level Topic Structure corpus (CPTS), achieving high quality. We also build several strong baselines, including ChatGPT, to validate the computability of CPTS on two fundamental tasks (topic segmentation and outline generation) and preliminarily verified its usefulness for the downstream task (discourse parsing).
 \\ \newline  }

\begin{document}

\maketitleabstract

\section{Introduction}
\begin{figure}[htbp]
\centerline{\includegraphics[width=7cm]{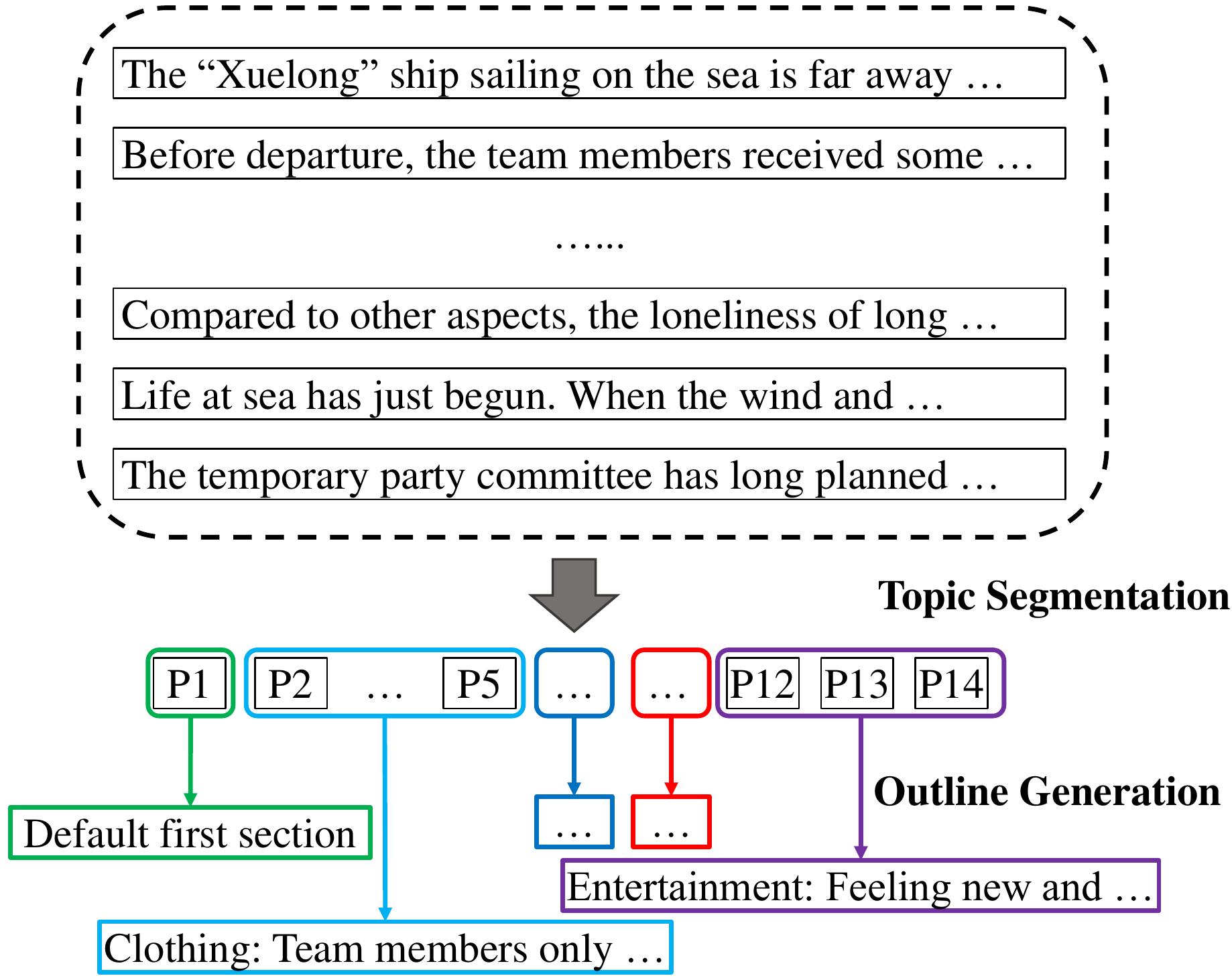}}
\caption{A document consisting of 14 basic units (paragraphs) is divided into five sections (different colors) according to the topic, and each of them has a subheading as its topic.}
\label{fig_example}
\end{figure}

A well-written document usually consists of several semantically coherent text segments, each of which revolves around a specific topic. Such topic structure can be discovered by topic segmentation and outline generation, which gives an overall grasp of the document. Topic segmentation aims to detect the segments (i.e., sentence or paragraph groups) in documents, and the subsequent task outline generation is to generate the corresponding subheading of each segment. Figure \ref{fig_example} shows an example of two tasks at the paragraph level where the basic units are paragraphs. 

Compared with sentence-level topic structure, the paragraph-level topic structure pays more attention to the document's higher-level topic structure between paragraphs, which can benefit quickly grasping and understanding the overall context of the document. It not only benefits traditional downstream NLP tasks, such as document summarization~\cite{xiao-carenini-2019-extractive} and discourse parsing~\cite{DBLP:conf/aaai/JiangFCLZ021,huber2021predicting}, but also play an important role in Large Language Model (LLM) era. For example, during Retrieval Augment Generation (RAG) for large language models, obtaining the required information from long documents is necessary. The paragraph-level topic structure of a document can help quickly locate the approximate location of the desired content in long documents, reducing search space. 

Thanks to the development of topic structure theory~\cite{van1983strategies,chafe1994discourse, Todd2003Topics, stede2011discourse} in English,  more and more work in English has focused on topic segmentation on realistic datasets since the first attempt on synthetic datasets~\citeplanguageresource{DBLP:conf/anlp/Choi00}. They are not only limited to high-quality manually annotated datasets~\citeplanguageresource{DBLP:conf/emnlp/EisensteinB08,DBLP:conf/naacl/ChenBBK09} but also large-scale datasets~\citeplanguageresource{DBLP:conf/naacl/KoshorekCMRB18,DBLP:journals/tacl/ArnoldSCGL19} automatically constructed from structured source data such as WIKI. The annotation content of topic structure has gradually enriched, from only topic boundaries to using words or phrases to annotate the topics of each text segment~\citeplanguageresource{liu-etal-2022-end}. 

However, there are fewer studies on Chinese topic structure compared to English. Most previous work focused on sentence-level topic segmentation in dialogues~\citeplanguageresource{DBLP:conf/aaai/XuZ021, zhang2023mug} or WIKI text~\citeplanguageresource{DBLP:conf/ijcnlp/XingHCT20} following English works. In the paragraph-level topic structure, ~\citetlanguageresource{DBLP:conf/nlpcc/WangLXL16} annotated 2951 web documents with paragraph-level topic structures, but the corpus is unfortunately inaccessible, and specific annotation details are not disclosed.

Therefore, it is necessary to construct a large-scale and high-quality paragraph-level topic structure corpus to serve the Chinese research community. Considering the success of English and the current shortage of resources in Chinese, there are two challenges in filling the gap in Chinese paragraph-level topic structure research. 

The first challenge is how to represent paragraph-level topic structures more richly. Most paragraph-level corpora (e.g., Cites and Elements corpora~\cite{DBLP:conf/naacl/ChenBBK09}) following sentence-level topic structure representation use keywords or phrases as topic contents~\cite{DBLP:conf/naacl/KoshorekCMRB18,DBLP:journals/tacl/ArnoldSCGL19}. Since basic units (paragraphs) are longer than the sentence level, using keywords or phrases cannot express richer topic information they contain~\cite{todd2016discourse}. It led subsequent studies to focus more on topic segmentation~\cite{DBLP:conf/anlp/Choi00,riedl2012topictiling,badjatiya2018attention,somasundaran2020two} and ignore outline generation. While some studies~\cite{DBLP:conf/sigir/ZhangGFLC19,DBLP:conf/acl/BarrowJMMOR20,DBLP:conf/emnlp/LoJTLDB21} have attempted to generate outlines, they only classify topics into limited types instead of generating real subheadings. 

Another challenge is how to build a paragraph-level topic structure corpus that is both large-scale and high-quality. On the one hand, the existing high-quality manual corpora~\cite{DBLP:conf/naacl/ChenBBK09, DBLP:conf/nlpcc/WangLXL16} are relatively small since the annotation requires assigning the topic attribution of paragraphs instead of sentences, which is laborious and time-consuming~\cite{seale1997ensuring,todd2011analyzing}. Besides, the manual annotation for topic content may be subjective and different from the author's intention due to topic ambiguity. On the other hand, although the automatic extraction method can build large-scale corpora~\cite{DBLP:conf/naacl/KoshorekCMRB18,DBLP:journals/tacl/ArnoldSCGL19}, it only ensures the correctness of the topic structure and content in terms of form, but not in terms of semantics that is crucial for outline generation.

To address the above issues, we first propose a hierarchical paragraph-level topic structure representation for modeling the topic structure of documents more comprehensively inspired by English topic theories. It has a three-layered topic structure, not only including paragraph and topic boundaries but also subheadings and the title. Especially, it takes the real subheading (clause or sentence) rather than keywords or phrases to represent topic content, ensuring the richness of the topic information longer basic units contain.

Then, we propose a two-stage man-machine collaborative annotation method to construct the Chinese Paragraph-level Topic Structure corpus (CPTS) with about 14393 documents with high quality based on our representation. Specifically, in the first stage, we first use a heuristics automatic extraction method for the topic boundary and content (subheading) from more common unstructured new documents, keeping the subject of the topic and the large scale of the corpus. In the second stage, to ensure the high quality of the corpus, we ask the human verifiers to verify the extracted topic structure instead of manually annotating them, which can greatly reduce the workload. Using this two-stage construction process of first extracting and then verifying, we build the largest Chinese paragraph-level topic structure corpus with a high quality (94.79\% Inter-Annotator Agreement and 0.849 Kappa value).

Finally, to verify the computability of the CPTS, we construct several strong baselines, including ChatGPT, on two basic tasks: topic segmentation and outline generation. Also, preliminary experiments in the downstream task discourse parsing have verified the usefulness of its topic structure.~\footnote{We release the corpus and baselines at \url{https://github.com/fjiangAI/CPTS}.}

\section{Related Work}
\label{sec:append-how-prod}

\subsection{Topic Structure Theory}
Unlike the intra-sentence topic structure, which is often a keyword, we focus on the discourse topic structure above sentences and paragraphs. \citet{van1983strategies} considered discourse topics as a series of super propositions consisting of sentence propositions. \citet{chafe1994discourse} stated that a discourse topic is a collection of relevant events, states, and references that agree in some form with the speaker's semi-active consciousness. \citet{Todd2003Topics} argued that discourse topics are clusters of similar or related concepts to create connectivity and relevance. 

Although different discourse topic theories have different views on the form of topics, researchers have roughly the same definition of discourse topic boundaries. ~\cite{stede2011discourse} pointed out that a document would consist of several topics, each containing one or more basic units describing the same topic. The length of a topic will vary depending on the length of the document or the purpose of the research~\cite{moens2001generic,ponte1997text}. At the paragraph level, ~\cite{DBLP:journals/coling/Hearst97} regarded the paragraph boundaries as the potential topic boundary.

\subsection{Topic Structure Corpus}
There are many corpora at the sentence level, including constructed by synthetic~\citeplanguageresource{DBLP:conf/anlp/Choi00}, manually annotated~\citeplanguageresource{DBLP:conf/emnlp/EisensteinB08}, and automatic extracted~\citeplanguageresource{DBLP:conf/naacl/KoshorekCMRB18,DBLP:journals/tacl/ArnoldSCGL19} method. In Chinese, following the schema in English, the Wiki section zh~\citeplanguageresource{DBLP:conf/ijcnlp/XingHCT20} is a sentence-level topic structure corpus containing 10K documents randomly selected from the Chinese Wikipedia. XZZ ~\citeplanguageresource{DBLP:conf/aaai/XuZ021} is manually annotated 505 recorded conversations, but it only annotated sentence-level topic boundaries and did not annotate topic contents. The subsequent MUG~\citeplanguageresource{zhang2023mug} supplemented it by annotating 654 conversations with sentence-level topic boundaries as well as the topic content.

However, due to the large basic units (paragraphs), the paragraph-level topic annotation is laborious and time-consuming~\cite{seale1997ensuring,todd2011analyzing}. The manually annotated corpora are relatively small, such as Cities and Elements~\citeplanguageresource{DBLP:conf/naacl/ChenBBK09}, which only have about 100 documents. Therefore, recent research has shifted towards automatic extraction~\citeplanguageresource{liu-etal-2022-end}. There is relatively little research in Chinese, and the manually annotated WLX~\citeplanguageresource{DBLP:conf/nlpcc/WangLXL16} is a paragraph-level topic structure corpus containing 2951 documents, but unfortunately, their dataset is not publicly accessible.

\subsection{Topic Segmentation and Outline Generation Method}
In English, early work mainly used unsupervised methods~\cite{DBLP:journals/coling/Hearst97,DBLP:conf/anlp/Choi00,riedl2012topictiling,DBLP:conf/starsem/GlavasNP16} for topic segmentation. Owing to having constructed the large-scale topic structure corpora, supervised methods have gradually become mainstream, such as the sequential labeling models~\cite{badjatiya2018attention,DBLP:conf/naacl/KoshorekCMRB18,somasundaran2020two, Lukasik2020}, and the pointer networks~\cite{Li2018}. Only a few studies focused on the Chinese topic segmentation task by following English methods using the sequential labeling models~\cite{DBLP:conf/nlpcc/WangLXL16,DBLP:conf/ijcnlp/XingHCT20} or local classification model~\cite{DBLP:conf/aaai/JiangFCLZ021} to predict the topic boundary.

Most of the previous corpora in English annotated topic contents as keywords or phrases, which are very short. It caused the outline generation works to be easily formulated as a classification problem in a joint learning framework with topic segmentation~\cite{DBLP:conf/sigir/ZhangGFLC19,DBLP:conf/acl/BarrowJMMOR20,DBLP:conf/emnlp/LoJTLDB21}. There are few research in Chinese due to a lack of suitable corpora. MUG~\citeplanguageresource{zhang2023mug} views topic segmentation and outline generation as two separate tasks for benchmarks.

\section{Chinese Paragraph-level Topic Structure Representation}

A proper topic structure representation is a prerequisite and necessary condition for guiding the construction of a topic corpus. It determines the form and content of corpus annotation. Most of the existing corpora only annotate basic units and topics they subordinate following \citet{goutsos1997modeling}'s theory. Recent work has gradually enriched annotations, such as using keywords and phrases to annotate topic content. However, at the paragraph level, as the granularity becomes larger, richer annotations are needed to comprehensively express the high-level structure of the document, such as subheadings and titles.

\begin{figure}[htbp]
\centerline{\includegraphics[width=7cm]{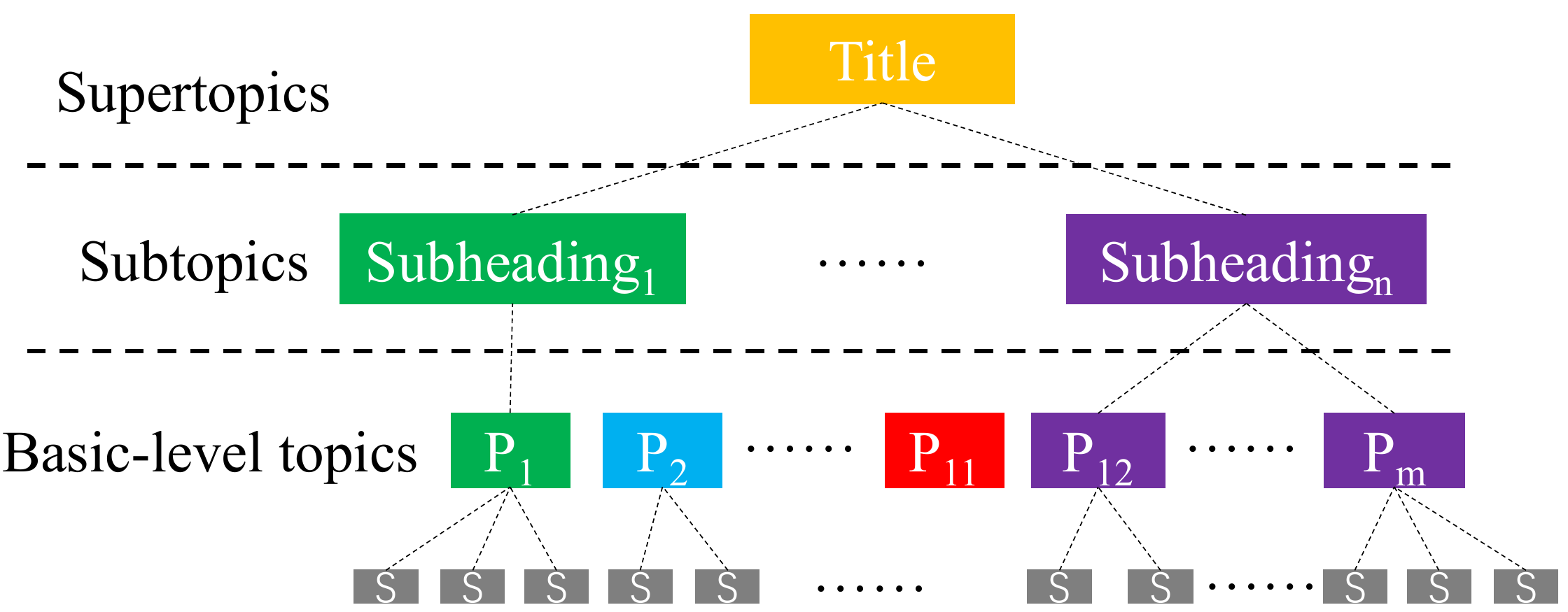}}
\caption{The Chinese paragraph-level topic structure representation, which contains $m$ paragraphs and $n$ subheadings. We assume a paragraph only has one topic, and a topic contains one or more paragraphs. S represents a sentence.}
\label{fig_framwork}
\end{figure}

\begin{figure*}[htbp]
\centerline{\includegraphics[width=15cm]{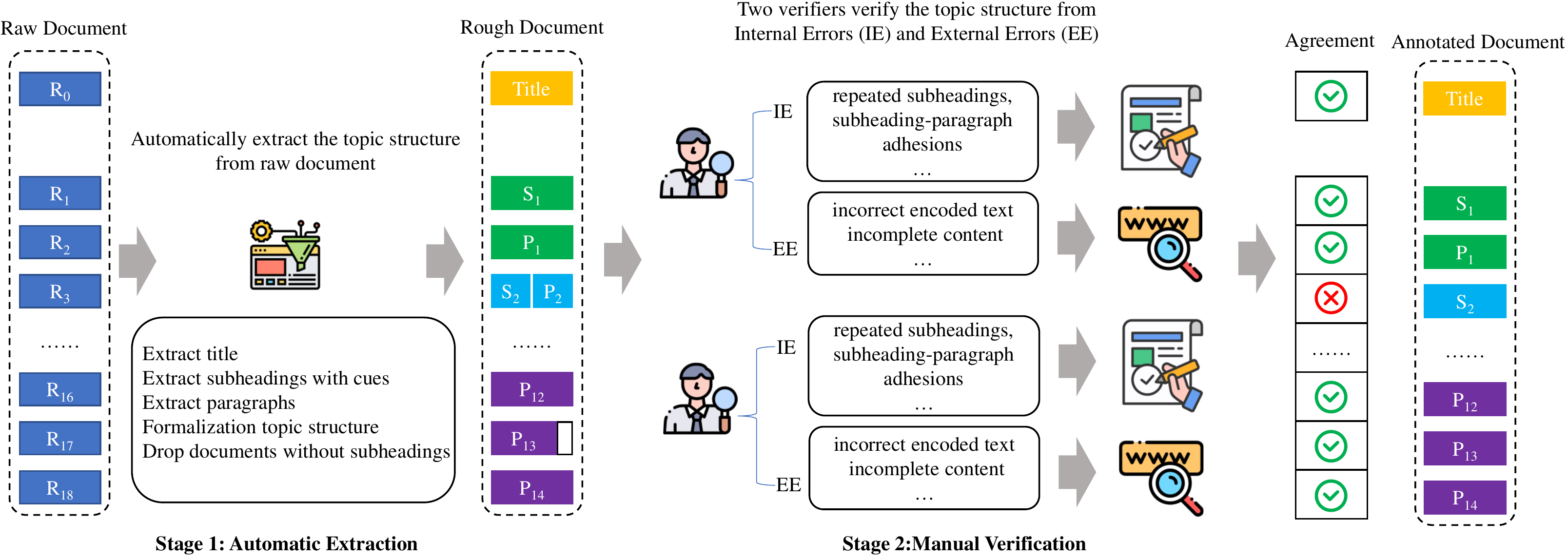}}
\caption{The two-stage man-machine collaborative annotation process. R$_n$ means raw paragraphs, S$_n$ means subheadings and P$_n$ means annotated paragraphs.} \label{fig_process}
\end{figure*}

Recognizing this, we propose a three-layer hierarchical representation of the Chinese paragraph-level topic structure for guiding corpus construction according to discourse topic theories~\cite{bruning1999cognitive,van2014discourse}: \emph{supertopics}, \emph{subtopics}, and \emph{basic level topics}, as illustrated in Figure~\ref{fig_framwork}. It not only includes paragraph boundaries and topic boundaries but also includes topic content and the higher-level title of a document. In particular, we regard the subheading and title as a clause or sentence instead of keywords to represent richer information on the paragraph level.

Specifically, we take the document's title as a supertopic, subheadings as subtopics, and paragraphs as basic-level topics, which implies a single paragraph will belong to one topic, and consecutive paragraphs describing the same topic fall under the same subheading. All subheadings, in turn, are subordinate to the title as subtopics. This hierarchical topic structure can capture relationships not only between paragraphs and subheadings but also between subheadings and the title.

In addition, we use subheadings (clauses or sentences) instead of noun phrases to represent subtopics of longer basic units (paragraphs), overcoming the limitation of keywords or phrases as subtopics that limit the amount of information conveyed by the topic at the paragraph level (the subtopic only has 4.7 tokens in latest work~\cite{liu-etal-2022-end}). By integrating the semantic richness of subheadings, we can better capture the nuances of the document's content and structure.

\section{Chinese Paragraph-level Topic Structure Corpus Construction}

\subsection{Data Source}
Although our proposed representation model can be applied to various genres of documents, we still want to construct the corresponding corpus from more general text to assist downstream tasks better. Therefore, we select the news documents issued by Xinhua News Agency from Chinese Gigaword Fourth Edition~\citeplanguageresource{ChineseGigawordFourthEdition} (Gigaword corpus) as the data source for generalization. It contains 1373448 news documents of four types (i.e., advis, multi, other and story) from January 1991 to December 2008. We chose story news as candidate documents because they account for most of them (1314198/1373448) and are more standardized than the other three types of news, which is conducive to building a more generalized topic structure.

\subsection{Man-machine Collaborative Annotation}
Manually constructing topic structure corpora is time-consuming and limited in scale due to topic ambiguities~\cite{seale1997ensuring,todd2011analyzing}, while large-scale corpora constructed by automatic extraction without manual verification do not guarantee the correctness of topic boundaries and content in semantics that are essential for finer tasks such as outline generation. Thus, inspired by previous work on automatic~\cite{DBLP:conf/naacl/KoshorekCMRB18} and manual construction~\cite{DBLP:conf/emnlp/EisensteinB08}, we design a two-stage man-machine collaborative annotation strategy involving both automatic extraction and manual verification to build a large-scale and high-quality corpus.

The details of our two-stage man-machine collaborative annotation process are shown in Figure~\ref{fig_process}. In the first stage, we take several steps to automatically extract candidate documents with topic structures, ensuring the correctness of topic boundaries and contents in terms of form. In the second stage, each document is double-checked for internal and external errors by two human validators to ensure the correctness of topic boundaries and contents in terms of semantics. 

\subsubsection{Automatic Extraction}

Following previous work~\cite{DBLP:conf/naacl/KoshorekCMRB18}, we automatically extracted the topic structure of the document in the first stage. Different from easily extracting topic structure from structured text in WIKI, news text is harder because it is typically composed of unstructured paragraphs with natural texts. Therefore, we designed a heuristic automatic extraction method to extract topic structures from raw documents automatically. Firstly, we extract the first paragraph that only includes one sentence and without ending punctuation as the title of a document. Then, we traverse the following paragraphs in the documents. If a paragraph has only one sentence and has a special token "(subheading)", we regard it as a subheading. Otherwise, it will be added to the paragraph list. After traversing all paragraphs, we formalize the topic structure representation based on the position of the paragraphs and subheadings and drop the document that does not contain any subheadings. This automatic extraction method can quickly extract topic structures from a large number of documents, but due to its simple heuristic rules, the accuracy cannot be guaranteed. Therefore, we take the second stage of manual verification to make it up.

\subsubsection{Manual Verification}
\citet{DBLP:journals/coling/Hearst97} pointed out that most documents do not contain explicit subheadings that indicate the topic structure. Therefore, after the first stage, we obtain 14393 (about 1\% of raw documents) rough documents containing subheadings extracted automatically since few documents explicitly have two or more subheadings with special   tokens in the Gigaword corpus. To ensure the correctness of topic boundaries and content in terms of semantics, we ask verifiers to verify the topic structure of each rough document. Our verification team consists of one Ph.D. student, six master's students, and one senior undergraduate student, all of whom are engaged in natural language processing. They are divided into four groups, and each document will be verified by one group (two verifiers) to ensure the objectivity and accuracy of verification. It is worth mentioning that since the sub-topics are automatically extracted in the first stage, the verifier simply needs to check the correctness of paragraphs, subheadings, and title rather than label boundaries and write topic contents, which significantly reduces the annotation effort. 

As shown in Figure~\ref{fig_process}, manual verification mainly verifies the correctness of automatic extraction from a semantic perspective at this stage and also quickly re-verifies the form correctness of topic boundaries that have been automatically extracted. For semantic issues, they are mainly checked from both internal and external errors. Internal errors refer to errors that verifiers can correct through the document itself, including repeated subheadings, title-paragraph adhesions, etc. For external errors such as incorrect encoded text or incomplete content in some subheadings or paragraphs that cannot be fixed from the document itself, verifiers use the help of search engines to retrieve the source news and make corresponding modifications.

During the second verification stage, two verifiers in each group verify the documents separately. When two verifiers have a conflict in their annotations, they discuss resolving. Finally, 36\% of documents containing errors are revised by our verifiers. Thanks to the automatic extraction of most of the correct topic structures in the first stage, the average Inter-Annotator Agreement (IAA) between two verifiers of the same group is 94.79\%, and the Kappa value~\cite{cohen1960coefficient} between them is 0.849. It demonstrates that our two-stage man-machine collaborative annotation method only requires the verifier to validate the topic structure rather than directly generate it, significantly reducing the workload while maintaining high quality. 

\begin{table}[htbp]
\centering
\resizebox{\linewidth}{!}
{
\begin{tabular}{lllr}
\hline
  \textbf{Item}                     & \textbf{Max} & \textbf{Min} & \textbf{Avg.}  \\ \hline
\# words/document & 5791  & 180   & 1727.96 \\
\# paragraphs/document & 40  & 2   & 14.76 \\
\# words/subheading & 147  & 1   & 12.33 \\
\# paragraphs/subheading    & 33  & 1   & 3.70  \\
\# subheadings/document     & 20  & 2   & 4.00  \\\hline
\end{tabular}
}
\caption{ The statistical details of CPTS.}
\label{table:1}
\end{table}

\begin{figure*}[htbp]
	\centering
	\begin{subfigure}[t]{5cm}
		\centering
		\includegraphics[width=5cm]{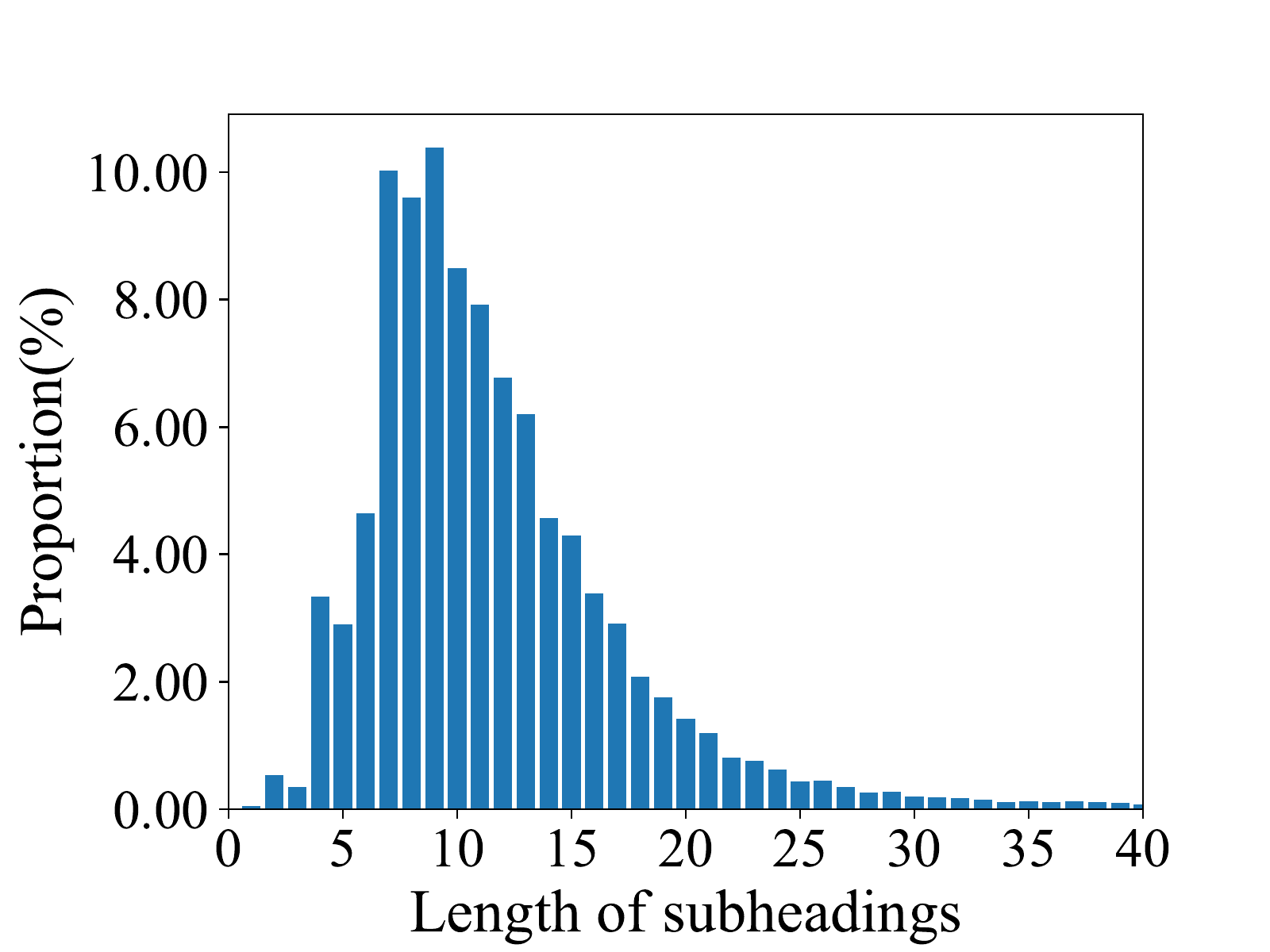}
		\subcaption{Distribution of subheadings length.}\label{fig_subheadings}
	\end{subfigure}
	\quad
	\begin{subfigure}[t]{5cm}
		\centering
		\includegraphics[width=5cm]{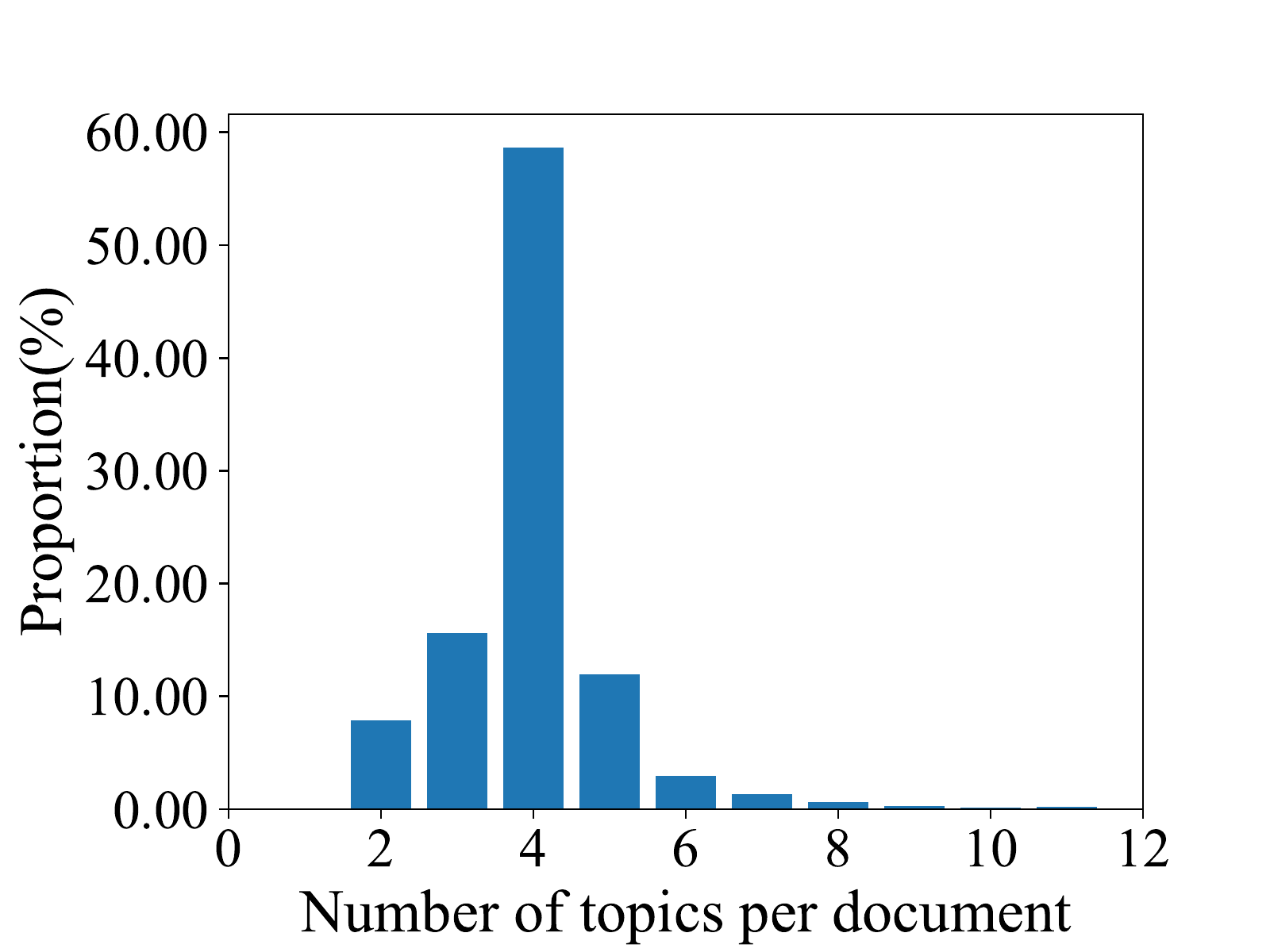}
		\subcaption{Distribution of topics per document.}\label{fig_topics}
	\end{subfigure}
	\quad
	\begin{subfigure}[t]{5cm}
		\centering
		\includegraphics[width=5cm]{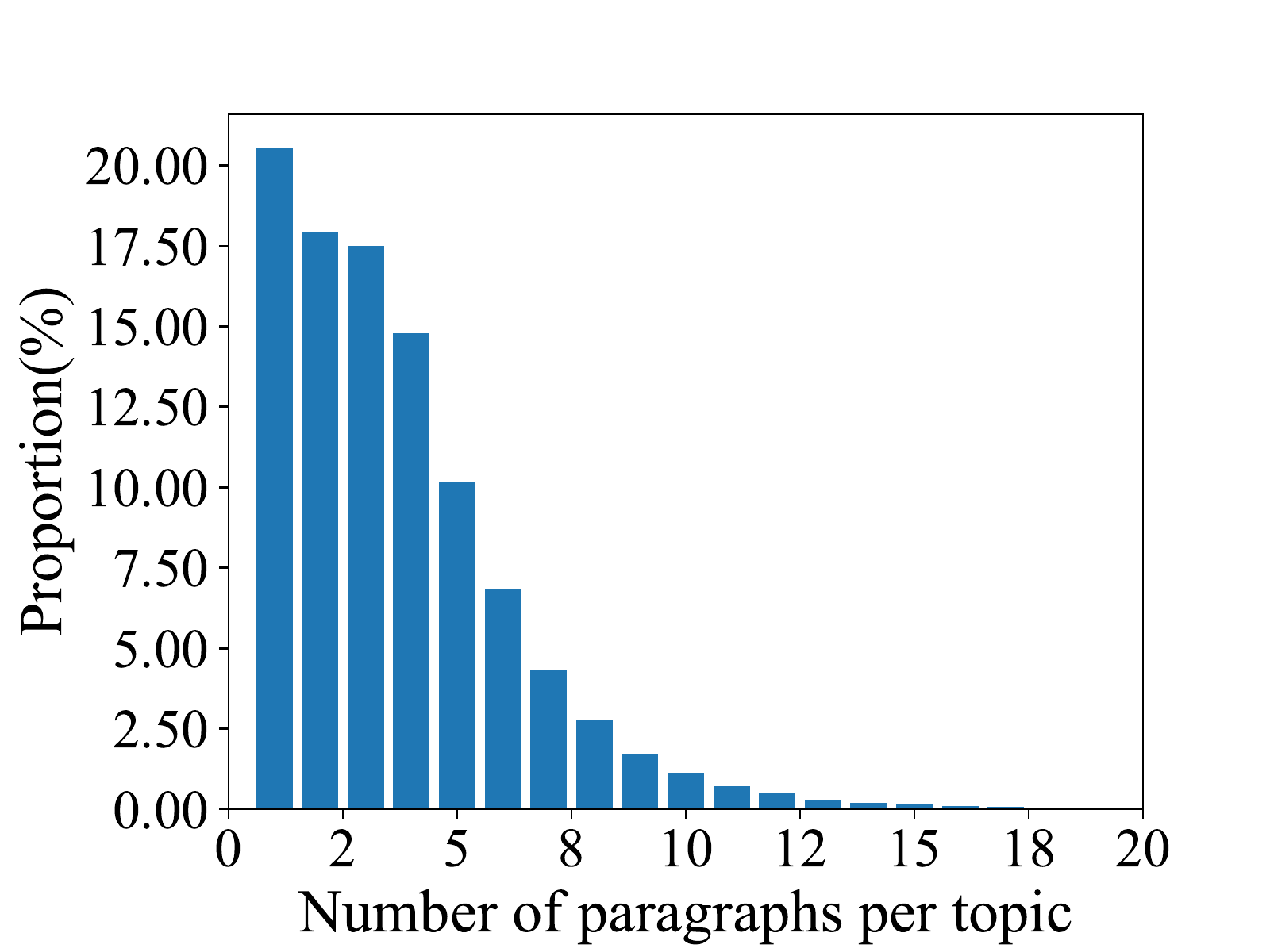}
		\subcaption{Distribution of paragraphs per topic.}\label{fig_paragraphs}
	\end{subfigure}
	\quad
	\caption{The main distribution of the length of subheadings, topics per document, and paragraphs per topic.}\label{fig:distribution}
\end{figure*}

\section{Statistics and Analysis on CPTS}
\subsection{Details of CPTS corpus}

The details of CPTS are shown in Table~\ref{table:1}, the range and average figures for various aspects like the number of words per document (ranging from 180 to 5791, with an average of 1727.96), paragraphs per document (ranging from 2 to 40, averaging at 14.76), words per subheading (averaging at 3.70) and subheadings per document (ranging from 2 to 20, with an average of 4.00). Furthermore, Figure~\ref{fig:distribution} depicts the main distributions of the length of subheadings, topics per document, and paragraphs per topic. Figure~\ref{fig_subheadings} shows that about 90\% subheadings have more than seven words, and only a few subheadings have less than four words. It shows that subheadings in CPTS are usually clauses or sentences rather than words or phrases~\citep{DBLP:journals/tacl/ArnoldSCGL19}, which could fully express the information of a paragraph-level topic.  Figure~\ref{fig_topics} shows that about 60\% of the documents have four topics, demonstrating the topic granularity will change with the document length~\cite{todd2016discourse}. We also notice that over 70\% of topics contain less than four paragraphs in Figure~\ref{fig_paragraphs}. They indicate the usefulness of the paragraph-level topic: A document can be divided into two more simple structures through paragraph-level topics (the discourse structure among paragraph-level topics and that in one topic).

\begin{table*}[htbp]
\centering
\resizebox{\linewidth}{!}
{
\begin{tabular}{lclllllll}
\hline
Dataset & Scale & Genre & Topic level & Topic Form & Annotation Method & Annotation content & Support Tasks & Accessible \\
\hline
XZZ & 505 & Dialogue & sentence & - & manual & TB & TS & $\surd$ \\
MUG & 654 & Dialogue & sentence & clause or sentence & manual & PB, TB, Subheadings, Title & TS, OG, TG & $\surd$ \\
Wiki$_{zh}$ & 10000 & Wikipedia & sentence & phrase & automatic & TB & TS & $\times$* \\
WLX & 2951 & Web doc & paragraph & unknown & manual & Unknown & TS & $\times$ \\ \hline
CPTS(Ours) & 14393 & News text & paragraph & clause or sentence & man-machine collaborative & PB, TB, Subheadings, Title & TS, OG, TG & $\surd$ \\
\hline
\end{tabular}
}
\caption{The comparison of CPTS and the other Chinese corpora. The asterisk* means that Wiki section zh (Wiki$_{zh}$) contains 10000 documents randomly selected from ZhWiki and is not directly available.TB means Topic Boundary, PB means Paragraph Boundary, TS means Topic Segmentation, OG means Outline Generation, and TG means Title Generation.}
\label{tab:comparing}
\end{table*}

\subsection{Compared with Other Chinese Topic Structure Corpora}
The comparison of CPTS and the other Chinese topic structure corpora (Wiki$_{zh}$~\cite{DBLP:conf/ijcnlp/XingHCT20}, XZZ~\cite{DBLP:conf/aaai/XuZ021}, MUG~\citeplanguageresource{zhang2023mug} and WLX~\cite{DBLP:conf/nlpcc/WangLXL16}) are shown in Table~\ref{tab:comparing}. Firstly, thanks to our two-stage human-machine collaborative annotation method, we have constructed the largest high-quality Chinese topic structure corpus. It is about four times larger than the existing largest paragraph-level one (WLX), and even larger than the largest sentence-level corpus (Wiki$_{zh}$) by automatic extraction. At the same time, incorporating manual verification after automatic extraction makes our corpus maintain the same high quality as the manually annotated corpus (94.79\% IAA and 0.849 Kappa value).

Secondly, compared with other written corpus, CPTS annotated more comprehensive topic structures based on our Chinese paragraph-level topic structure representation, including Paragraph Boundaries (PB), Topic Boundaries (TB), subheadings, and titles, which benefit more tasks like Outline Generation (OG) and Title Generation (TG). 

Finally, CPTS will be open access to the community to fill the gap in the Chinese paragraph-level topic segmentation resources. It complements the sentence-level dialogue topic structure corpus mutually and fully supports the relevant tasks, including paragraph-level topic segmentation and outline generation, and even speeds up the retrieval of needed information from documents in the RAG process of LLM.

We also noticed the similarities between our CPTS and MUG, but there are two aspects of differences. Firstly, the annotated genre and topic level are different. We focus on paragraph-level topic structures in written texts, while MUG focuses on sentence-level topic structures in spoken dialogues. Secondly, the source of annotating content is different.  In our CPTS, paragraph boundaries, topic boundaries, subheadings, and the title are manually verified after being automatically extracted from the original document, which is closer to the author's intention. However, in the MUG, annotators manually write these from the reader aspect.

\section{Experiments on Corpus Evaluation}
To verify the computability of our annotated CPTS, we select several strong baselines to experiment on two basic tasks (i.e., topic segmentation and outline generation) of CPST as benchmarks for further research. Following previous work~\citeplanguageresource{zhang2023mug}, we build the baselines on these two tasks separately to obtain the objective and absolute performance of them. We randomly divided the dataset into training (90\%) and testing sets (10\%) according to the paragraph length distribution of the document for topic segmentation and outline generation. Specifically, there are 12953 documents in the training set and 1440 documents in the test set.

\subsection{Topic Segmentation}
\subsubsection{Task Definition and Baselines}
Following \citet{DBLP:conf/naacl/KoshorekCMRB18}, we view topic segmentation as a supervised learning task, as shown in Eq.~\ref{equation1}. The input $P$ is a document, represented as a sequence of $n$ paragraphs $(p_1,...,p_n)$, and the label $Y$ = $(y_{1},...,y_{n-1})$ is a segmentation of the document, represented by $n-1$ binary values, where $y_i$ denotes whether $p_i$ ends a segment. $Model_{ts}$ is a topic segmentation model.

\begin{equation}
    (y_{1},...,y_{n-1})=Model_{ts}(p_1,...,p_n)
    \label{equation1}
\end{equation}

For the topic segmentation task, we select the following three representative kinds of models as baselines. \textbf{Segbot}~\cite{Li2018} and \textbf{PN-XLNet} (a variant of Segbot where using XLNet replaces the GRU encoder) are two pointer network models that first encode input text by GRU or XLNet and then use a pointer network to select topic boundaries in the input sequence.  \textbf{TM-BERT}~\cite{DBLP:conf/aaai/JiangFCLZ021} is local classification model that identifies the topic boundary by a  triple semantic BERT-based matching mechanism. \textbf{BERT+Bi-LSTM} and \textbf{Hier.~BERT}~\cite{Lukasik2020} are sequential labeling models, using LSTM and Transformer as base-architecture, separately. Following the published papers, we reproduce them, and the experimental settings of each model in topic segmentation are shown in Table \ref{tab:ts_setting}.
We also take \textbf{ChatGPT} as a baseline and adopt the prompt and settings from the probing ChatGPT in conversation topic segmentation~\cite{fan2023uncovering}. 
\begin{table}[htbp]
\centering
\resizebox{\linewidth}{!}
{ 
\begin{tabular}{lllll}
\hline
\textbf{Model}         & \textbf{BS} & \textbf{LR} & \textbf{Epoch} & \textbf{PLM} \\ \hline
Segbot        & 20           & 1E-03        & 10    & Word2Vec             \\
PN-XLNet & 20           & 1E-03        & 10    & XLNet-base           \\\hline
TM-BERT       & 2            & 1E-05        & 10    & Bert-base            \\\hline
Bert+Bi-LSTM  & 2            & 1E-05        & 10    & Bert-base            \\
Hier.~BERT     & 2            & 1E-05        & 10    & Bert-base            \\ \hline
\end{tabular}
}
\caption{The main hyper-parameters of baselines on topic segmentation. BS is the batch size, LR is learning rate and PLM is the pre-trained language model.}
\label{tab:ts_setting}
\end{table}

\subsubsection{Evaluation and Results}
For topic segmentation evaluation, we use the following commonly used metrics\footnote{https://github.com/cfournie/segmentation.evaluation}: $P_k$, WindowDiff, Segmentation Similarity and Boundary Similarity. We also report the macro-F1 of each model for a comprehensive evaluation.

\begin{table}[htbp]
\centering
\resizebox{\linewidth}{!}
{ 
\begin{tabular}{lllllll}
\hline
\textbf{Model}   & \textbf{$P_k$↓} & \textbf{WD↓} & \textbf{S↑}    & \textbf{B↑}    & \textbf{F1↑} \\ \hline
ChatGPT (0-shot) & 41.12  & 63.57  & 37.45 & 59.51  & 52.51
\\ \hline
Segbot & 24.06  & 25.85  & 89.73 & 58.94  & 75.23    \\
PN-XLNet & 22.02  & 23.34  & 91.27 & 65.19  & 77.70    \\\hline
TM-BERT & 22.86  & 24.44  & 89.93 & 58.84  & 80.62 \\\hline
BERT+Bi-LSTM & \textbf{19.45}  & \textbf{20.89}  & 91.76 & 65.88  & \textbf{81.62}   \\
Hier.~BERT & 19.76  & 21.00  & \textbf{91.92} & \textbf{66.54}  & 81.40   \\\hline
\end{tabular}
}
\caption{The performance on topic segmentation. Different from Segmentation Simlarity (S) and Boundary Similarity (B), $P_k$ and WindowDiff (WD) are penalty measures.}
\label{tab:performance}

\end{table}

The experimental results are shown in Table~\ref{tab:performance}. Although ChatGPT is a powerful LLM, its performance in topic segmentation on text still lags far behind other fine-tuned pre-trained models due to 0-shot setting. Compared to Segbot without a pre-trained model, PN-XLNet improves the performance in all metrics. As a local classification model, TM-BERT outperforms PN-XLNet by 2.92 in F1 value, however, it performs worse in the popular topic segmentation evaluation matrix ($P k$, WD, S, and B). Besides, with the two-layer architecture, BERT+Bi-LSTM and Hier.~BERT achieve the best performance.  Both models outperform other models in terms of F1 values and other metrics, as BERT better captures the semantic representation of paragraphs and hierarchical modeling better captures global information.

\subsection{Outline Generation}
\subsubsection{Task Definition and Baselines}
Unlike the previous work~\cite{DBLP:conf/acl/BarrowJMMOR20} on outline generation that takes only the first-level heading (usually a word or a phrase) of the document in Wikipedia as its subheading, the subheading in CPTS is usually a clause or sentence. It is more challenging to joint learning of topic segmentation and outline generation~\cite{DBLP:conf/sigir/ZhangGFLC19}. Therefore, we align with MUG~\citeplanguageresource{zhang2023mug} and treat outline generation as a separate task like summary generation instead of text classification~\cite{DBLP:conf/acl/BarrowJMMOR20,DBLP:conf/emnlp/LoJTLDB21} to simplify the problem and achieve a more intuitive performance. Given a section $s_j$ that contains $m$ consecutive paragraphs $(p_j^1,...,p_j^m)$ in one topic, the outline generation model ($Model_{og}$) needs to generate the subheading $h_j$ of them, as shown in Eq.~\ref{equation2}.

\begin{equation}
    h_j=Model_{og}(p_j^1,...,p_j^m)
    \label{equation2}
\end{equation}

We select the following popular text generation models as our baselines. \textbf{BART}~\cite{DBLP:conf/acl/LewisLGGMLSZ20} and \textbf{T5}~\cite{DBLP:journals/jmlr/RaffelSRLNMZLL20} are popular encoder-decoder-based pre-trained models. All of them have the same settings: Batch-size is 8, LR is 1E-05 and epoch is 10. We also take the \textbf{ChatGPT} as a baseline since the strong generation ability (The corresponding prompt can be seen in Appendix~\ref{app:prompt}).

\subsubsection{Evaluation and Results}

Since outline generation is regarded as a summarization task, we evaluate baseline models using the evaluation methods commonly used in summarization. Specifically, we use ROUGE~\cite{lin-2004-rouge} and BLEU~\cite{papineni-etal-2002-bleu} to evaluate the quality of generated subheadings from the word overlap, and use BertScore~\cite{DBLP:conf/iclr/ZhangKWWA20} to evaluate the quality of generated subheadings from the semantics. Furthermore, we also did a manual evaluation by ranking their results on 100 randomly selected samples. The manual evaluation details can be seen in Appendix~\ref{app:manual_eval}.

\begin{table}[htbp]
\centering
\resizebox{\linewidth}{!}
{ 
\begin{tabular}{lrrrrcc}
\hline
\textbf{Model} & \textbf{R-1} & \textbf{R-2} & \textbf{R-L} & \textbf{BLEU}  & \textbf{BertScore} & \textbf{Rank ↓} \\ \hline
ChatGPT (0-shot)   & 22.64  & 12.04  & 20.58 & 6.49  & 61.39 & \textbf{2.49}    \\
ChatGPT (3-shot) & 22.25  &11.87  &20.22  &6.47  &61.45 & 2.61 \\\hline
BART& 25.86  & 16.20  & 24.50 & 12.55  & 63.49 & 3.68    \\ 
T5   & 27.14  & 16.00  & 25.44  & 12.04 & 63.74 & 3.25    \\
T5 (24)    & \textbf{28.91}  & \textbf{17.88}  & \textbf{27.06}  & \textbf{14.46} & \textbf{64.67} & 2.98     \\
\hline
\end{tabular}
}
\caption{The performance on outline generation.}
\label{tab:performance in outline generation}
\end{table}

Table~\ref{tab:performance in outline generation} shows the performance of baselines in outline generation. Compared to the poor performance in topic segmentation, ChatGPT's 0-shot performance is close to the other popular fine-tuned models in outline generation due to its strong text generation ability. However, the 3-shot setting does not achieve further significant improvement. One reason may be that three long texts as input affect the in-context-learning (ICL) capability of ChatGPT. BART and T5 achieve similar performance with the same scale (the 6-layer encoder and the 6-layer decoder). In Table~\ref{tab:performance in outline generation}, the T5 model outperforms BART on R-1, R-L, and BertScore by 1.28, 0.94, and 0.25, respectively, while BART performs better on R-2 and BLEU. Not surprisingly, the larger T5 model performs best among all baselines.

It is worth noting that subheadings are more refined, shorter, and more abstract than summaries. Therefore, even the performance of the SOTA model is still not ideal in evaluation (e.g., the ROUGE in outline generation is lower than that in the summarization task.). It shows that outline generation is challenging, and there is still room for improvement in future work.

We observe two interesting trends from the last right column that is the manual evaluation. On the one hand, ChatGPT (0-shot) and ChatGPT (3-shot), which are models that are not fine-tuned, perform better (average 2.49 and 2.61 separately) than the other three fine-tuned models on manual ranking. It indicates that ChatGPT-generated subheadings are more friendly for humans, even though they may not align precisely with the original subheadings. On the other hand, the fine-tuned models (BART, T5, and T5 (24)) show the consistency of orders between manual ranking and their respective Rouge, BLEU, and BertScore metrics evaluations. It shows the reliability of our automatic evaluation and the usefulness of our corpus.

\subsubsection{Title Generation Results}

Since we have proposed a three-level topic structure representation, we further added title generation as a supplement to outline generation. In this task, we use all subtitles as input to generate the title.

\begin{table}[htbp]
\centering
\resizebox{\linewidth}{!}
{ 
\begin{tabular}{lrrrrc}
\hline
\textbf{Model} & \textbf{R-1} & \textbf{R-2} & \textbf{R-L} & \textbf{BLEU}  & \textbf{BertScore} \\ \hline
ChatGPT(0-shot) & 16.87  & 7.79   & 15.08 & 3.85 & 59.52 
\\
ChatGPT(3-shot) & 16.81  & 7.60  & 15.00  & 3.69  & 59.31\\
\hline
BART& 25.85  & \textbf{16.62}  & 24.67 & \textbf{11.86}  & 63.79     \\ 
T5   & 25.06  & 14.19  & 23.47  & 8.86 & 62.76     \\
T5 (24)    & \textbf{28.01}  & 16.55   & \textbf{26.11}  & 10.96 & \textbf{64.61}     \\
\hline
\end{tabular}
}
\caption{The performance on title generation.}
\label{tab:performance in title generation}
\end{table}

Table~\ref{tab:performance in title generation} shows the performance of each model on the title generation task. The performance of each model is similar to that in the outline generation task. The lower performance of ChatGPT on title generation shows it is more challenging than outline generation since it needs more abstraction and reasoning at a higher level. It also demonstrates that outline and title generation are harder than traditional text summarization.

\section{Application in Discourse Parsing}
To validate the effectiveness of our constructed representation and corpus, we also utilized it to assist in discourse parsing, a downstream task. Different from topic segmentation, which splits the documents into several segments, discourse parsing is a more complex task where the model needs to build a tree structure on basic discourse units. Previous researches~\cite{DBLP:conf/aaai/JiangFCLZ021,huber2021predicting} have shown that topic structure can imply a skeleton of the rhetorical structure of a document, and there is a consistency in local discrimination for topic segmentation and discourse parsing. However, the lack of a topic structure corpus limits the method application in Chinese. Therefore, we attempt to use CPTS for training the model to predict topic structure in the document without explicit subheadings, helping discourse parsing. 

\begin{table}[htbp]
\centering
\begin{tabular}{lrrrrc}
\hline
\textbf{Model} & \textbf{Span} \\ \hline
Dist(Paragraph Boundary)& 50.23   \\ 
Dist(Topic Boundary) & 55.33      \\
\hline
\end{tabular}
\caption{The performance on MCDTB.}
\label{tab:performance on MCDTB2}
\end{table}

Specifically, we conducted our preliminary experiment on MCDTB~\cite{jiang2018mcdtb}, which is a popular corpus for discourse parsing in Chinese. Then, following the previous work~\cite{huber2021predicting}, we employed two distant supervision methods for paragraph-level discourse parsing. The Dist(Paragraph Boundary) model utilizes paragraph boundaries from the MCDTB corpus as topic boundaries to learn the discourse structure of the document, while the Dist(topic Boundary) model leverages the real topic structure provided by CPTS as the learning goal to learn the discourse structure of the document. Table~\ref{tab:performance on MCDTB2} shows that the real topic structure in CPTS enhances the parser's performance on paragraph-level discourse parsing in Chinese from 50.23 to 55.33, demonstrating the usefulness of the topic structure corpus.

\section{Discussion and Future Work}

\subsection{The Applicability of Annotation Method}
Our methodology is not limited to the news documents used in this paper and is equally applicable to other genres as long as they possess some markers indicating paragraph-level topic structures. This includes a wide range of textual materials such as legal documents, novels, and academic documents, where structured topics are often revealed by special tokens. Then, in our method, the annotator only needs to verify the internal and external errors after the automatic extraction with the special token, reducing the annotation workload greatly.

A particularly promising application is in the analysis of scripts or novels, where our method could be used to identify broader narrative shifts or stage changes indicated by unique transition and voiceover tokens in the script. We believe that by constructing paragraph-level topic structures for these diverse categories, our method will not only aid in a deeper understanding and grasp of the paragraph-level topic structure of documents but also could guide large language models (LLMs) in generating more controlled content at the higher level according to given structures.

\subsection{The Potential Challenges}

\textbf{Expansion of the Joint Learning Framework of Topic Segmentation and Outline Generation}. There are some works attempting to jointly learn topic segmentation and outline generation by two classification tasks due to shorter subheadings. Expanding the current joint learning framework is a viable approach to deal with longer subheading challenges. A possible solution is integrating heterogeneous text classification tasks (such as topic segmentation) into text generation tasks (like outline generation) into a single and unified generation model based on powerful LLMs. 

\textbf{Exploration of Hierarchical Topic Structures}. Existing methods usually view the topic structure as a flat structure and ignore the longer dependency of different topics. Therefore, another promising direction is delving into the hierarchical nature of paragraph-level topic structures. This exploration can be conducted from both bottom-up and top-down perspectives. By harnessing the capabilities of large-scale language models, it's possible to model the hierarchical relationships among topics (including parent-child connections between different layers) and the interconnections between various topic segments within the same document, such as writing style and semantic relationships with more words.

\section{Conclusion}

To fill the gap in Chinese paragraph-level topic structure resources, we first propose a three-layer discourse topic structure representation to guide the construction of our corpus. It takes the sentence as the topic to express richer paragraph-level information and incorporates paragraph boundaries, topic boundaries, subheadings, and the title into the topic structure. Then, we designed a two-stage human-machine collaborative annotation method to construct the largest high-quality Chinese paragraph-level topic structure corpus. By combining automatic extraction and manual verification, we ensure the correctness of the topic structure not only formally but also semantically. We described the construction process of the corpus in detail and conducted an in-depth analysis and comparison of it. Finally, we verified the computability of the corpus through eight topic segmentation and outline generation baselines, including ChatGPT. In the future, we will focus on improving the performance of Chinese topic segmentation and outline generation by designing appropriate methods to assist other downstream tasks in the LLM era.

\section{Acknowledgements}
This research is supported by the project of Shenzhen Science and Technology Research Fund (Fundamental Research Key Project Grant No. JCYJ20220818103001002), the Internal Project Fund from Shenzhen Research Institute of Big Data under Grant No. T00120-220002, Shenzhen Key Laboratory of Cross-Modal Cognitive Computing (grant number ZDSYS20230626091302006), and the National Natural Science Foundation of China (No. 62376181).

\nocite{*}
\section{Bibliographical References}\label{sec:reference}

\bibliographystyle{lrec-coling2024-natbib}
\bibliography{lrec-coling2024-example}

\begin{thebibliography}{60}
\expandafter\ifx\csname natexlab\endcsname\relax\def\natexlab#1{#1}\fi

\bibitem[{Arnold et~al.(2019)Arnold, Schneider, Cudr{\'{e}}{-}Mauroux, Gers, and L{\"{o}}ser}]{DBLP:journals/tacl/ArnoldSCGL19}
Sebastian Arnold and Rudolf Schneider and Philippe Cudr{\'{e}}{-}Mauroux and Felix A. Gers and Alexander L{\"{o}}ser. 2019.
\newblock \emph{{SECTOR:} {A} Neural Model for Coherent Topic Segmentation and Classification}.

\bibitem[{Badjatiya et~al.(2018)Badjatiya, Kurisinkel, Gupta, and Varma}]{badjatiya2018attention}
Pinkesh Badjatiya, Litton~J Kurisinkel, Manish Gupta, and Vasudeva Varma. 2018.
\newblock Attention-based neural text segmentation.
\newblock In \emph{ECIR}, pages 180--193.

\bibitem[{Barrow et~al.(2020)Barrow, Jain, Morariu, Manjunatha, Oard, and Resnik}]{DBLP:conf/acl/BarrowJMMOR20}
Joe Barrow, Rajiv Jain, Vlad~I. Morariu, Varun Manjunatha, Douglas~W. Oard, and Philip Resnik. 2020.
\newblock A joint model for document segmentation and segment labeling.
\newblock In \emph{ACL}, pages 313--322.

\bibitem[{Bruning et~al.(1999)Bruning, Schraw, and Ronning}]{bruning1999cognitive}
Roger~H Bruning, Gregory~J Schraw, and Royce~R Ronning. 1999.
\newblock \emph{Cognitive psychology and instruction}.
\newblock ERIC.

\bibitem[{Chafe(1994)}]{chafe1994discourse}
Wallace Chafe. 1994.
\newblock \emph{Discourse, consciousness, and time: The flow and displacement of conscious experience in speaking and writing}.

\bibitem[{Chen et~al.(2009)Chen, Branavan, Barzilay, and Karger}]{DBLP:conf/naacl/ChenBBK09}
Harr Chen and S. R. K. Branavan and Regina Barzilay and David R. Karger. 2009.
\newblock \emph{Global Models of Document Structure using Latent Permutations}.

\bibitem[{Choi(2000)}]{DBLP:conf/anlp/Choi00}
Freddy Y. Y. Choi. 2000.
\newblock \emph{Advances in domain independent linear text segmentation}.

\bibitem[{Choi et~al.(2001)Choi, Wiemer-Hastings, and Moore}]{choi2001latent}
Freddy~YY Choi, Peter Wiemer-Hastings, and Johanna~D Moore. 2001.
\newblock Latent semantic analysis for text segmentation.
\newblock In \emph{EMNLP}.

\bibitem[{Choubey and Huang(2021)}]{Choubey2021}
Prafulla~Kumar Choubey and Ruihong Huang. 2021.
\newblock {Profiling News Discourse Structure Using Explicit Subtopic Structures Guided Critics}.
\newblock In \emph{Findings of EMNLP}, pages 1594--1605.

\bibitem[{Cohen(1960)}]{cohen1960coefficient}
Jacob Cohen. 1960.
\newblock A coefficient of agreement for nominal scales.
\newblock \emph{Educational and psychological measurement}, 20(1):37--46.

\bibitem[{Devlin et~al.(2019)Devlin, Chang, Lee, and Toutanova}]{Devlin2019bert}
Jacob Devlin, Ming-Wei Chang, Kenton Lee, and Kristina Toutanova. 2019.
\newblock Bert: Pre-training of deep bidirectional transformers for language understanding.
\newblock In \emph{NAACL-HLT}, pages 4171--4186.

\bibitem[{Eisenstein(2009)}]{DBLP:conf/naacl/Eisenstein09}
Jacob Eisenstein. 2009.
\newblock Hierarchical text segmentation from multi-scale lexical cohesion.
\newblock In \emph{NAACL-HLT}, pages 353--361.

\bibitem[{Eisenstein and Barzilay(2008)}]{DBLP:conf/emnlp/EisensteinB08}
Jacob Eisenstein and Regina Barzilay. 2008.
\newblock \emph{Bayesian Unsupervised Topic Segmentation}.

\bibitem[{Fan et~al.(2023)Fan, Jiang, Li, and Li}]{fan2023uncovering}
Yaxin Fan, Feng Jiang, Peifeng Li, and Haizhou Li. 2023.
\newblock \href {http://arxiv.org/abs/2305.08391} {Uncovering the potential of chatgpt for discourse analysis in dialogue: An empirical study}.

\bibitem[{Fournier(2013)}]{fournier2013evaluating}
Chris Fournier. 2013.
\newblock Evaluating text segmentation using boundary edit distance.
\newblock In \emph{ACL}, pages 1702--1712.

\bibitem[{Fournier and Inkpen(2012)}]{fournier2012segmentation}
Chris Fournier and Diana Inkpen. 2012.
\newblock Segmentation similarity and agreement.
\newblock In \emph{NAACL-HLT}, pages 152--161.

\bibitem[{Glavas et~al.(2016)Glavas, Nanni, and Ponzetto}]{DBLP:conf/starsem/GlavasNP16}
Goran Glavas, Federico Nanni, and Simone~Paolo Ponzetto. 2016.
\newblock Unsupervised text segmentation using semantic relatedness graphs.
\newblock In \emph{*SEM@ACL}.

\bibitem[{Glavas and Somasundaran(2020)}]{somasundaran2020two}
Goran Glavas and Swapna Somasundaran. 2020.
\newblock Two-level transformer and auxiliary coherence modeling for improved text segmentation.
\newblock In \emph{AAAI}, pages 7797--7804.

\bibitem[{Goutsos(1997)}]{goutsos1997modeling}
Dionysis Goutsos. 1997.
\newblock \emph{Modeling discourse topic: sequential relations and strategies in expository text}, volume~59.
\newblock Greenwood Publishing Group.

\bibitem[{Grimes(2015)}]{grimes2015thread}
Joseph~E Grimes. 2015.
\newblock \emph{The thread of discourse}.
\newblock De Gruyter Mouton.

\bibitem[{Hearst(1997)}]{DBLP:journals/coling/Hearst97}
Marti~A. Hearst. 1997.
\newblock Texttiling: Segmenting text into multi-paragraph subtopic passages.
\newblock \emph{Comput. Linguistics}, 23(1):33--64.

\bibitem[{Huber et~al.(2022{\natexlab{a}})Huber, Xing, and Carenini}]{huber2021predicting}
Patrick Huber, Linzi Xing, and Giuseppe Carenini. 2022{\natexlab{a}}.
\newblock Predicting above-sentence discourse structure using distant supervision from topic segmentation.
\newblock In \emph{AAAI}.

\bibitem[{Huber et~al.(2022{\natexlab{b}})Huber, Xing, and Carenini}]{huber2022predicting}
Patrick Huber, Linzi Xing, and Giuseppe Carenini. 2022{\natexlab{b}}.
\newblock Predicting above-sentence discourse structure using distant supervision from topic segmentation.
\newblock In \emph{AAAI}.

\bibitem[{Jiang et~al.(2021)Jiang, Fan, Chu, Li, Zhu, and Kong}]{DBLP:conf/aaai/JiangFCLZ021}
Feng Jiang, Yaxin Fan, Xiaomin Chu, Peifeng Li, Qiaoming Zhu, and Fang Kong. 2021.
\newblock Hierarchical macro discourse parsing based on topic segmentation.
\newblock In \emph{AAAI}, pages 13152--13160.

\bibitem[{Jiang et~al.(2018)Jiang, Xu, Chu, Li, Zhu, and Zhou}]{jiang2018mcdtb}
Feng Jiang, Sheng Xu, Xiaomin Chu, Peifeng Li, Qiaoming Zhu, and Guodong Zhou. 2018.
\newblock {MCDTB}: A macro-level {C}hinese discourse {T}ree{B}ank.
\newblock In \emph{COLING}, pages 3493--3504.

\bibitem[{Koshorek et~al.(2018)Koshorek, Cohen, Mor, Rotman, and Berant}]{DBLP:conf/naacl/KoshorekCMRB18}
Omri Koshorek and Adir Cohen and Noam Mor and Michael Rotman and Jonathan Berant. 2018.
\newblock \emph{Text Segmentation as a Supervised Learning Task}.

\bibitem[{Lewis et~al.(2020)Lewis, Liu, Goyal, Ghazvininejad, Mohamed, Levy, Stoyanov, and Zettlemoyer}]{DBLP:conf/acl/LewisLGGMLSZ20}
Mike Lewis, Yinhan Liu, Naman Goyal, Marjan Ghazvininejad, Abdelrahman Mohamed, Omer Levy, Veselin Stoyanov, and Luke Zettlemoyer. 2020.
\newblock {BART:} denoising sequence-to-sequence pre-training for natural language generation, translation, and comprehension.
\newblock In \emph{ACL}, pages 7871--7880.

\bibitem[{Li et~al.(2018)Li, Sun, and Joty}]{Li2018}
Jing Li, Aixin Sun, and Shafiq Joty. 2018.
\newblock {SEGBOT: A generic neural text segmentation model with pointer network}.
\newblock In \emph{IJCAI}, pages 4166--4172.

\bibitem[{Lin(2004{\natexlab{a}})}]{lin-2004-rouge}
Chin-Yew Lin. 2004{\natexlab{a}}.
\newblock \href {https://aclanthology.org/W04-1013} {{ROUGE}: A package for automatic evaluation of summaries}.
\newblock In \emph{Text Summarization Branches Out}, pages 74--81, Barcelona, Spain. Association for Computational Linguistics.

\bibitem[{Lin(2004{\natexlab{b}})}]{lin2004rouge}
Chin-Yew Lin. 2004{\natexlab{b}}.
\newblock Rouge: A package for automatic evaluation of summaries.
\newblock In \emph{Text summarization branches out}, pages 74--81.

\bibitem[{Liu and Lapata(2019)}]{DBLP:conf/emnlp/LiuL19}
Yang Liu and Mirella Lapata. 2019.
\newblock Text summarization with pretrained encoders.
\newblock In \emph{EMNLP-IJCNLP}, pages 3728--3738.

\bibitem[{Liu et~al.(2022)Liu, Zhu, and Zeng}]{liu-etal-2022-end}
Liu, Yang and Zhu, Chenguang and Zeng, Michael. 2022.
\newblock \href {https://doi.org/10.18653/v1/2022.findings-acl.46} {\emph{End-to-End Segmentation-based News Summarization}}.
\newblock Association for Computational Linguistics.

\bibitem[{Lo et~al.(2021)Lo, Jin, Tan, Liu, Du, and Buntine}]{DBLP:conf/emnlp/LoJTLDB21}
Kelvin Lo, Yuan Jin, Weicong Tan, Ming Liu, Lan Du, and Wray~L. Buntine. 2021.
\newblock Transformer over pre-trained transformer for neural text segmentation with enhanced topic coherence.
\newblock In \emph{Findings of EMNLP}, pages 3334--3340.

\bibitem[{Lukasik et~al.(2020)Lukasik, Dadachev, Papineni, and Sim{\~{o}}es}]{Lukasik2020}
Michal Lukasik, Boris Dadachev, Kishore Papineni, and Gon{\c{c}}alo Sim{\~{o}}es. 2020.
\newblock {Text Segmentation by Cross Segment Attention}.
\newblock In \emph{EMNLP}, pages 4707--4716.

\bibitem[{Malioutov and Barzilay(2006)}]{DBLP:conf/acl/MalioutovB06}
Igor Malioutov and Regina Barzilay. 2006.
\newblock Minimum cut model for spoken lecture segmentation.
\newblock In \emph{ACL}.

\bibitem[{Moens and De~Busser(2001)}]{moens2001generic}
Marie-Francine Moens and Rik De~Busser. 2001.
\newblock Generic topic segmentation of document texts.
\newblock In \emph{SIGIR}, pages 418--419.

\bibitem[{Papineni et~al.(2002)Papineni, Roukos, Ward, and Zhu}]{papineni-etal-2002-bleu}
Kishore Papineni, Salim Roukos, Todd Ward, and Wei-Jing Zhu. 2002.
\newblock \href {https://doi.org/10.3115/1073083.1073135} {{B}leu: a method for automatic evaluation of machine translation}.
\newblock In \emph{ACL}, pages 311--318, Philadelphia, Pennsylvania, USA. Association for Computational Linguistics.

\bibitem[{Pevzner and Hearst(2002)}]{pevzner2002critique}
Lev Pevzner and Marti~A Hearst. 2002.
\newblock A critique and improvement of an evaluation metric for text segmentation.
\newblock \emph{Computational Linguistics}, 28(1):19--36.

\bibitem[{Ponte and Croft(1997)}]{ponte1997text}
Jay~M Ponte and W~Bruce Croft. 1997.
\newblock Text segmentation by topic.
\newblock In \emph{International Conference on Theory and Practice of Digital Libraries}, pages 113--125.

\bibitem[{Radford et~al.(2019)Radford, Wu, Child, Luan, Amodei, Sutskever et~al.}]{radford2019language}
Alec Radford, Jeffrey Wu, Rewon Child, David Luan, Dario Amodei, Ilya Sutskever, et~al. 2019.
\newblock Language models are unsupervised multitask learners.
\newblock \emph{OpenAI blog}, 1(8):9.

\bibitem[{Raffel et~al.(2020)Raffel, Shazeer, Roberts, Lee, Narang, Matena, Zhou, Li, and Liu}]{DBLP:journals/jmlr/RaffelSRLNMZLL20}
Colin Raffel, Noam Shazeer, Adam Roberts, Katherine Lee, Sharan Narang, Michael Matena, Yanqi Zhou, Wei Li, and Peter~J. Liu. 2020.
\newblock Exploring the limits of transfer learning with a unified text-to-text transformer.
\newblock \emph{J. Mach. Learn. Res.}, 21:140:1--140:67.

\bibitem[{Riedl and Biemann(2012)}]{riedl2012topictiling}
Martin Riedl and Chris Biemann. 2012.
\newblock Topictiling: a text segmentation algorithm based on lda.
\newblock In \emph{Proceedings of ACL 2012 Student Research Workshop}, pages 37--42.

\bibitem[{{Robert Parker, David Graff, Ke Chen, Junbo Kong, Kazuaki Maeda}(2009)}]{ChineseGigawordFourthEdition}
{Robert Parker, David Graff, Ke Chen, Junbo Kong, Kazuaki Maeda}. 2009.
\newblock \emph{Chinese Gigaword Fourth Edition}.
\newblock Linguistic Data Consortium. ISLRN \href{https://www.islrn.org/resources/261-416-300-929-8}{261-416-300-929-8}.

\bibitem[{Seale and Silverman(1997)}]{seale1997ensuring}
Clive Seale and David Silverman. 1997.
\newblock Ensuring rigour in qualitative research.
\newblock \emph{The European journal of public health}, 7(4):379--384.

\bibitem[{Stede(2011)}]{stede2011discourse}
Manfred Stede. 2011.
\newblock Discourse processing.
\newblock \emph{Synthesis Lectures on Human Language Technologies}, 4(3):1--165.

\bibitem[{Todd(2011)}]{todd2011analyzing}
Richard~Watson Todd. 2011.
\newblock Analyzing discourse topics and topic keywords.

\bibitem[{Todd(2016)}]{todd2016discourse}
Richard~Watson Todd. 2016.
\newblock \emph{Discourse topics}, volume 269.
\newblock John Benjamins Publishing Company.

\bibitem[{Utiyama and Isahara(2001)}]{DBLP:conf/acl/UtiyamaI01}
Masao Utiyama and Hitoshi Isahara. 2001.
\newblock A statistical model for domain-independent text segmentation.
\newblock In \emph{ACL}, pages 491--498.

\bibitem[{Van~Dijk(2014)}]{van2014discourse}
Teun~A Van~Dijk. 2014.
\newblock \emph{Discourse and knowledge: A sociocognitive approach}.
\newblock Cambridge University Press.

\bibitem[{Van~Dijk and Kintsch(1983)}]{van1983strategies}
Teun~A Van~Dijk and Walter Kintsch. 1983.
\newblock \emph{Strategies of discourse comprehension}.
\newblock Acadamic Press.

\bibitem[{Wang et~al.(2016)Wang, Li, Xiao, and Lyu}]{DBLP:conf/nlpcc/WangLXL16}
Liang Wang and Sujian Li and Xinyan Xiao and Yajuan Lyu. 2016.
\newblock \emph{Topic Segmentation of Web Documents with Automatic Cue Phrase Identification and {BLSTM-CNN}}.

\bibitem[{Watson~Todd(2003)}]{Todd2003Topics}
Richard Watson~Todd. 2003.
\newblock \emph{Topics in classroom discourse}.
\newblock Ph.D. thesis, UK: University of Liverpool.

\bibitem[{Xiao and Carenini(2019)}]{xiao-carenini-2019-extractive}
Wen Xiao and Giuseppe Carenini. 2019.
\newblock \href {https://doi.org/10.18653/v1/D19-1298} {Extractive summarization of long documents by combining global and local context}.
\newblock In \emph{EMNLP-IJCNLP}, pages 3011--3021, Hong Kong, China. Association for Computational Linguistics.

\bibitem[{Xing et~al.(2020)Xing, Hackinen, Carenini, and Trebbi}]{DBLP:conf/ijcnlp/XingHCT20}
Linzi Xing and Brad Hackinen and Giuseppe Carenini and Francesco Trebbi. 2020.
\newblock \emph{Improving Context Modeling in Neural Topic Segmentation}.

\bibitem[{Xing et~al.(2022)Xing, Huber, and Carenini}]{xing2022improving}
Linzi Xing, Patrick Huber, and Giuseppe Carenini. 2022.
\newblock Improving topic segmentation by injecting discourse dependencies.
\newblock In \emph{Proceedings of 3rd Workshop on Computational Approaches to Discourse (CODI 2022)}, page~7.

\bibitem[{Xu et~al.(2021)Xu, Zhao, and Zhang}]{DBLP:conf/aaai/XuZ021}
Yi Xu and Hai Zhao and Zhuosheng Zhang. 2021.
\newblock \emph{Topic-Aware Multi-turn Dialogue Modeling}.

\bibitem[{Yang et~al.(2019)Yang, Dai, Yang, Carbonell, Salakhutdinov, and Le}]{yang2019xlnet}
Zhilin Yang, Zihang Dai, Yiming Yang, Jaime Carbonell, Russ~R Salakhutdinov, and Quoc~V Le. 2019.
\newblock Xlnet: Generalized autoregressive pretraining for language understanding.
\newblock In \emph{NeurIPS}.

\bibitem[{Zhang et~al.(2023)Zhang, Deng, Liu, Yu, Chen, Wang, Yan, Liu, Ren, and Zhao}]{zhang2023mug}
Zhang, Qinglin and Deng, Chong and Liu, Jiaqing and Yu, Hai and Chen, Qian and Wang, Wen and Yan, Zhijie and Liu, Jinglin and Ren, Yi and Zhao, Zhou. 2023.
\newblock \emph{Mug: A general meeting understanding and generation benchmark}.
\newblock IEEE.

\bibitem[{Zhang et~al.(2019)Zhang, Guo, Fan, Lan, and Cheng}]{DBLP:conf/sigir/ZhangGFLC19}
Ruqing Zhang, Jiafeng Guo, Yixing Fan, Yanyan Lan, and Xueqi Cheng. 2019.
\newblock Outline generation: Understanding the inherent content structure of documents.
\newblock In \emph{SIGIR}, pages 745--754.

\bibitem[{Zhang et~al.(2020)Zhang, Kishore, Wu, Weinberger, and Artzi}]{DBLP:conf/iclr/ZhangKWWA20}
Tianyi Zhang, Varsha Kishore, Felix Wu, Kilian~Q. Weinberger, and Yoav Artzi. 2020.
\newblock Bertscore: Evaluating text generation with {BERT}.
\newblock In \emph{ICLR}.

\end{thebibliography}

\appendix

\section{Appendix}
\subsection{The prompt we designed for outline generation}
\label{app:prompt}
The prompt we designed for the 0-shot setting in the outline generation is the following: 

\begin{CJK}{UTF8}{gbsn}
instruction:"一个文章包含几个段落(paragraph)和一个小标题(subheading)，请根据下面的段落生成文章的小标题，填写在subheading属性中，并以json的格式返回。" (A document contains several paragraphs and a subheading. Please generate a subheading for the document based on the following paragraphs, fill it in the subheading attribute, and return it in JSON format.)

input: <sample>
\end{CJK}

where <sample> is a dictionary containing several paragraphs on one topic: 
"input": "\{"paragraph": [...], "subheading": ""\}"

The prompt for the 3-shot setting is similar to the above.

\subsection{The details of manual evaluation on outline generation task}\label{app:manual_eval}

We randomly selected 100 samples and asked three evaluators to rank them based on the following evaluation settings.

In the manual evaluation, evaluators were provided with a text fragment with several paragraphs belonging to a single topic as the prompt. Subsequently, they were presented with output results from the five models as ranking candidates. Importantly, these options were initially randomized and anonymized to ensure an unbiased evaluation process.
The evaluators were asked to rank these candidate subheadings (from 1 to 5, 1 is the best and 5 is the worst) based on the following three criteria:

- Relevance: Whether the subheading accurately represents the content described in the text.

- Appropriateness: Whether the subheading conforms to typical styles and formats of subheadings. 

- Fluency: Whether the subheading is correctly formulated and flows smoothly. 

We then compiled the average rankings for each model from three evaluators to determine the final scores, ensuring the objectivity of our evaluation.

\end{document}